\documentclass[10pt,twocolumn,letterpaper]{article}

\usepackage[pagenumbers]{wacv} 

\usepackage{graphicx}
\usepackage{amsmath}
\usepackage{amssymb}
\usepackage{booktabs}
\usepackage{tabularx}
\usepackage{cellspace}
\usepackage{multirow}
\usepackage{tabularray} 
\usepackage{svg}
\usepackage[Export]{adjustbox}
\usepackage{pgfplots}
\usepackage[accsupp]{axessibility}  

\usepackage[pagebackref,breaklinks,colorlinks]{hyperref}

\usepackage[capitalize]{cleveref}
\crefname{section}{Sec.}{Secs.}
\Crefname{section}{Section}{Sections}
\Crefname{table}{Table}{Tables}
\crefname{table}{Tab.}{Tabs.}

\def\wacvPaperID{548} 
\def\confName{WACV}
\def\confYear{2024}

\begin{document}

\title{Improving Fairness using Vision-Language Driven Image Augmentation}


\author{
Moreno D'Inc\`a$^{1}$ \and 
Christos Tzelepis$^{2}$ \and 
Ioannis Patras$^2$ \and 
Nicu Sebe$^1$ \\ \and
$^1$University of Trento\\
{\tt\small moreno.dinca@unitn.it, niculae.sebe@unitn.it}
\and
$^2$Queen Mary University of London\\
{\tt\small \{c.tzelepis, i.patras\}@qmul.ac.uk}
}

\maketitle

\begin{abstract}
   Fairness is crucial when training a deep-learning discriminative model, especially in the facial domain. Models tend to correlate specific characteristics (such as \textit{age} and \textit{skin color}) with unrelated attributes (downstream tasks), resulting in biases which do not correspond to reality. It is common knowledge that these correlations are present in the data and are then transferred to the models during training (e.g.,~\cite{Torralba_2011_CVPR}). This paper proposes a method to mitigate these correlations to improve fairness. To do so, we learn interpretable and meaningful paths lying in the semantic space of a pre-trained diffusion model (DiffAE)~\cite{preechakul2021diffusion} -- such paths being supervised by contrastive text dipoles. That is, we learn to edit protected characteristics (\textit{age} and \textit{skin color}). These paths are then applied to augment images to improve the fairness of a given dataset. We test the proposed method on CelebA-HQ and UTKFace on several downstream tasks with \textit{age} and \textit{skin color} as protected characteristics. As a proxy for fairness, we compute the difference in accuracy with respect to the protected characteristics. Quantitative results show how the augmented images help the model improve the overall accuracy, the aforementioned metric, and the disparity of equal opportunity. Code is available at: \url{https://github.com/Moreno98/Vision-Language-Bias-Control}. \vspace{-0.3cm}
\end{abstract}

\section{Introduction}

    \begin{figure}[t]
        \centering
        \includegraphics[width=0.7\linewidth]{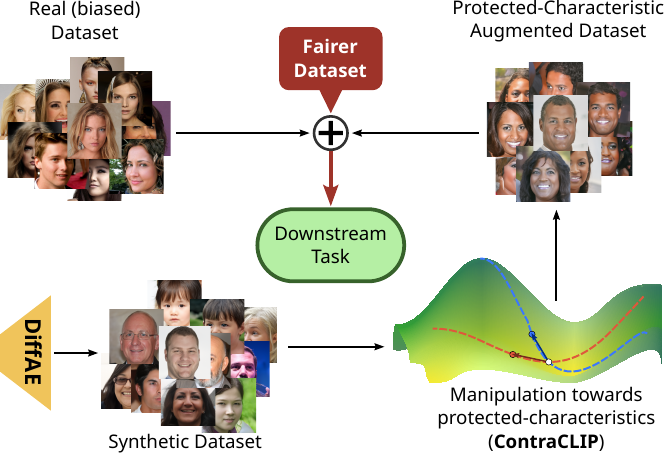}
        \caption{We make a biased dataset fairer by augmenting it with images (generated by DiffAE~\cite{preechakul2021diffusion}) depicting the desired protected characteristics (e.g., dark-skinned people) after being manipulated by our ContraCLIP~\cite{tzelepis2022contraclip}-based text-driven augmentation module.}
        \label{fig:teaser}
        \vspace{-0.4cm} 
    \end{figure}
    
    Today's society is careful about ethical topics and with the raising of publicly available AI tools~\cite{ramesh2022hierarchical, rombach2021highresolution, Karras_2019_CVPR} concerns about their fairness are also growing. In a supervised learning setting, the importance of the training data is well-known since the behavior of the model at inference time is highly correlated to the seen data. Modern models can effectively learn and highly perform multiple downstream tasks generalizing to unseen data. Besides the effectiveness of the pipeline, training data also brings unwanted side effects. It has been proven that vision datasets contain biases~\cite{Torralba_2011_CVPR}, thus the models learn the correlations present in the data which may be malignant~\cite{bolukbasi2016man, zhao2017men, Hendricks_2018_ECCV, pmlr-v81-buolamwini18a}. These concerns become a particularly sensitive subject when it comes to the facial domain. Modern machine learning models are dominant at a wide range of applications, such as face/emotion recognition and mask detection~\cite{Liu_2018_CVPR, Karras_2019_CVPR, d2023unleashing}. In this context, studying the behavior of deep learning models is crucial to avoid unwanted situations at inference time~\cite{Paleyes_2022}. For example, the model's performance may drop when presented with a particular protected characteristic (e.g., very young/old or dark-skinned faces). The above issues motivate us to study the behavior of a deep learning model with respect to facial protected characteristics which are sensitive to society and can raise ethical concerns.

    Training fair discriminative models has become of paramount importance for the research community during the past years. Recent works have shown that not only do models learn the underlying bias present in the data~\cite{Stock_2018_ECCV, jung2022learning}, but they tend to often amplify it~\cite{Hendricks_2018_ECCV, wang2020towards, zhao2017men}. Multiple techniques have been proposed for mitigating the bias, from task-specific training, such as the introduction of regularization terms or architectural approaches~\cite{wang2020towards, nam2020learning, intra-processing-debiasing-2020} to data augmentation strategies~\cite{Agarwal_2022_WACV, 9866115}. 
    
    Recently, generative models, such as Generative Adversarial Networks (GANs)~\cite{goodfellow2020generative}, have shown remarkable performance in a multitude of tasks through discovering controllable generative paths in their latent or feature spaces~\cite{oldfield2023panda,oldfield2021tensor,tzelepis2021warpedganspace,bounareli2023stylemask,bounareli2023iccv}. Thus, GAN-based methods have been employed as a data augmentation technique to generate fairer data~\cite{xu2018fairgan,chaudhari2022fairgen,xu2023ffpdg}, to generate counterfactuals~\cite{Dash_2022_WACV, abroshan2022counterfactual} or to generate counterparts by editing sensitive attributes~\cite{zhang2023fairnessaware}. The above works train generative models from scratch which may be impractical, especially in low data regimes. Additionally, the pre-trained generative models are expected to reflect the bias that is inherent to the datasets where they have been trained on~\cite{zhang2023fairnessaware,xu2018fairgan,oldfield2023part}, challenging those methods that use them for bias mitigation.
    
    In this paper, we address the above limitations by proposing a novel approach that leverages a \textit{pre-trained} diffusion model~\cite{preechakul2021diffusion,song2020denoising} to edit sensitive attributes in facial images, in order to improve the fairness of existing (biased) datasets and, consequently, the fairness of a discriminative model trained on such datasets. By contrast to previous works that train generative models (e.g., GANs~\cite{goodfellow2020generative}) from scratch~\cite{xu2018fairgan, Dash_2022_WACV, zhang2023fairnessaware}, we incorporate the power of a fixed pre-trained diffusion model to change sensitive facial attributes from a pool of generated images. The manipulated faces (with respect to the desired sensitive attributes) are used to make the original dataset fairer and mitigate the bias present on a downstream model trained on the original dataset. We illustrate this in Fig.~\ref{fig:teaser}. Our setting consists of a binary downstream classification attribute and a binary sensitive attribute towards which the model may exhibit bias. Throughout the paper, we will be referring to the downstream classification attribute as \textit{attribute} and to the sensitive attribute as \textit{protected characteristic}. The main contributions of this paper can be summarized as follows:
    \vspace{-0.2cm}
    \begin{itemize}
        \item To the best of our knowledge, we are the first to leverage diffusion models and a vision-language model in the context of fairness on discriminative models.
        \vspace{-0.2cm}
        \item We propose to edit the protected characteristics by learning interpretable paths in the semantic space of DiffAE~\cite{preechakul2021diffusion} guided by natural language without the need to fine-tune the \textit{pre-trained} generative model.\
        \vspace{-0.2cm}
        \item We test our method on several downstream tasks of CelebA-HQ~\cite{liu2015faceattributes} and UTKFace~\cite{zhifei2017cvpr} with \textit{skin color} and \textit{age} as protected characteristics showing competitive or better performance when mitigating the bias.\vspace{-0.2cm}
        \item We show that our method is capable of increasing the bias towards a specific attribute if needed -- that is, in contrast to previous works (e.g.,~\cite{nam2020learning,zhang2023fairnessaware,wang2020towards}), our method can control (i.e., decrease or increase) the bias concerning a specific attribute.
    \end{itemize}
    \vspace{-0.2cm}

\begin{figure*}[t]
    \centering
    \includegraphics[width=0.85\textwidth]{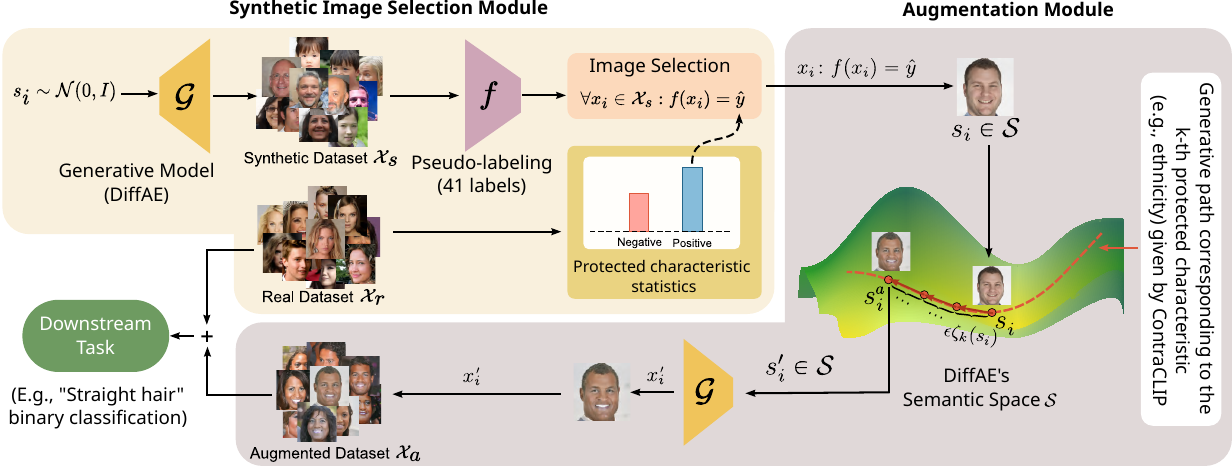}
    \caption{Overview of the proposed Vision-Language Bias Control (VLBC) method for controlling the bias in facial image datasets. Given a real training set $\mathcal{X}_r$ and a downstream task, we find the under-represented protected characteristic (e.g., black in \textit{skin colour}) by computing the sample statistics. Based on this, we select which images from a synthetic dataset $\mathcal{X}_s$ (generated using the DiffAE~\cite{preechakul2021diffusion} generator $\mathcal{G}$ and pseudo-labelled by a pre-trained network on 41 attributes $f$~\cite{liu2015faceattributes,karkkainenfairface}) to use for augmentation. Then, the selected images are manipulated by our augmentation module (ContraCLIP+DiffAE), pre-trained on text prompts defining the desired protected characteristic. In this example, we manipulate/augment the selected images based on \textit{skin colour}. Note that the original labels of the augmented images (i.e., corresponding to the attribute class at hand) do not change. Finally, the augmented dataset $\mathcal{X}_a$ is used along with the original real dataset $\mathcal{X}_r$ for training downstream tasks.}
    \label{fig:overview}
    \vspace*{-0.3cm}
\end{figure*}

\section{Related work}\label{sec:related_work}
    
    During recent years, the research community has directed its efforts towards mitigating bias and improving fairness in discriminative models mainly adopting one of the following approaches: i) proposing training techniques without changing the training data at hand or ii) applying some sort of data augmentation for the under-represented classes. We review each category below. \vspace{-0.4cm} 
    \paragraph{Training techniques} Different strategies have been investigated when mitigating the bias at training time, from the employment of regularization terms in the loss~\cite{nam2020learning} to architectural methods~\cite{wang2020towards, intra-processing-debiasing-2020}. Wang et al.~\cite{wang2020towards} proposed a new benchmark for bias mitigation by studying various techniques, such as oversampling rare examples, adversarial training and domain discriminative training, and proposing independent domain training where the downstream task is independently learnt by leveraging two independent heads, one for each domain class, leading to a model that is aware of the sensitive domain. Nam et al.~\cite{nam2020learning} proposed Learning from Failure (LfF), which simultaneously trains two classifiers, one to be biased and the other to be unbiased, by focusing on the hard samples for the biased one. This setting works under the assumption that a malignant bias is learnt when the sensitive attribute is easier to learn than the downstream target attribute; thus, if a biased model is struggling in learning a set of samples, then those samples will help in unbiasing a second model. This is achieved via generalized cross-entropy for amplifying the bias on the first model and via a weighted cross-entropy loss with relative difficulty for the target unbiased model. Savani et al.~\cite{intra-processing-debiasing-2020} introduce three intra-processing methods for bias mitigation, namely random perturbation where at each training iteration the model's weights are multiplied by Gaussian noise, layer-wise optimization where each layer of the network is optimized separately with a common objective function, and adversarial training where the model's bias is predicted via a trainable critic and used to improve the fairness avoiding the non-differentiable issue of bias metrics. \vspace{-0.4cm} 
    
    \paragraph{Generative data augmentation} A certain line of research proposes the generation of ``fairer'' data using generative models towards improving fairness. One of the first works in this direction is~\cite{xu2018fairgan}, where Sattigeri et al. introduce Fairness GAN, a GAN~\cite{goodfellow2020generative} conditioned on the protected sensitive attribute. This work generates a fairer dataset by training the proposed architecture on the original dataset. Dash et al.~\cite{Dash_2022_WACV} introduced a model trained to generate counterfactual versions of the same image based on the knowledge from a pre-defined causal graph. The synthesized images are then used to mitigate the bias by adding a regularization term to the loss, minimizing the MSE between the logits of the model when presented by the original image and the counterfactual. A drawback of this work is that the prior knowledge that is required from the causal graph for encoding the attribute and protected characteristic relations may not be readily available in practice. Zhang et al.~\cite{zhang2023fairnessaware} proposed a new setting where sensitive labels and downstream attributes are partially annotated. This work first learns a generator and a classifier with the available annotations, while subsequently the two models are incrementally trained using the mutual outputs. This semi-supervised setting of~\cite{zhang2023fairnessaware} is necessary due to the lack of annotations needed by the generator during training. The study also introduces a contrastive learning framework with balanced augmented data; i.e., for each original image, its counterpart is generated by changing the sensitive attribute, and then the downstream model is presented with both versions of the image pushing together images with different sensitive attributes and pushing away images with same sensitive attribute. Similarly, Jung et al.~\cite{jung2022learning} proposed the training of a model to pseudo-label a dataset with a specific sensitive attribute. Then, the pseudo-labels are filtered and replaced with random choices if the prediction confidence is lower than a certain threshold. Finally, the labeled and pseudo-labeled data are used to train a fairer model employing a training technique for bias mitigation (e.g., FairHSIC \cite{Quadrianto_2019_CVPR}). 
    
    The above generative techniques achieve notable bias mitigation results, but they typically require the training of domain-related generative models, demanding a significant amount of domain-specific data that may not be available for particular sensitive attributes because of their scarcity during data collection. Other works mitigate this problem via semi-supervised training~\cite{zhang2023fairnessaware}. In this work, we argue that a better solution exists. Since training a generative model on the available data may lead to a biased generator which could make the generation of rare examples challenging~\cite{xu2018fairgan}, we propose to explicitly learn to edit sensitive attributes by exploiting a \textit{pre-trained} generative model without the need of training or fine-tuning it.

    \begin{figure*}[t]
        \setlength\tabcolsep{0pt}
        \adjustboxset{width=0.70\linewidth,valign=c}
        \scriptsize
        \centering
        \begin{tabularx}{0.65\linewidth}{@{}
          X
          X
          X
          X
          X
          X
          X
          X
          X
          X
          X
        @{}}
        \multicolumn{11}{c}{\includegraphics{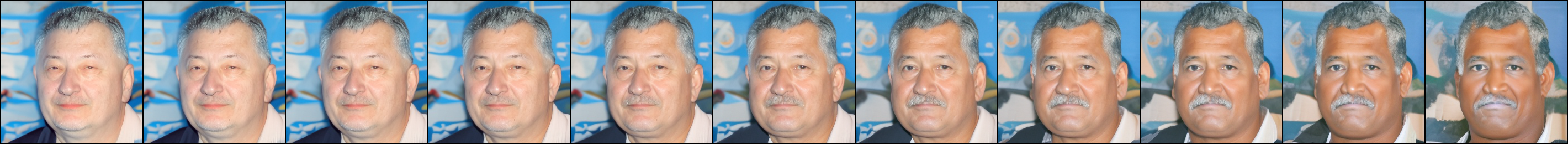}} \\
        \multicolumn{11}{c}{\includegraphics{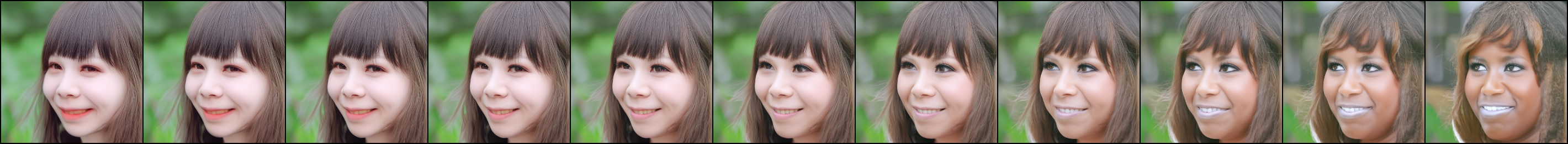}} \\
        \multicolumn{2}{c}{$\Leftarrow$ White} & & & & \multicolumn{1}{c}{Original} & & & & \multicolumn{2}{c}{Black $\Rightarrow$} \\
        \multicolumn{11}{c}{\includegraphics{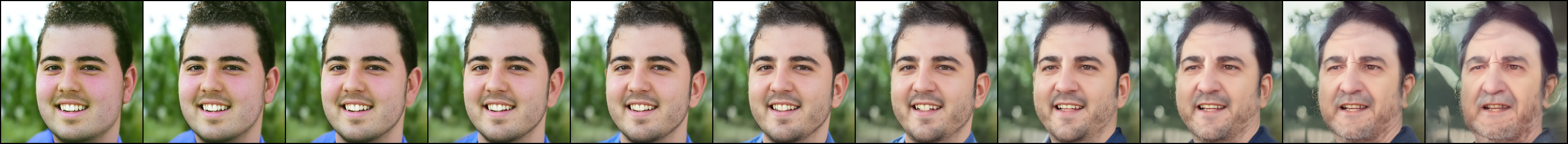}} \\
        \multicolumn{11}{c}{\includegraphics{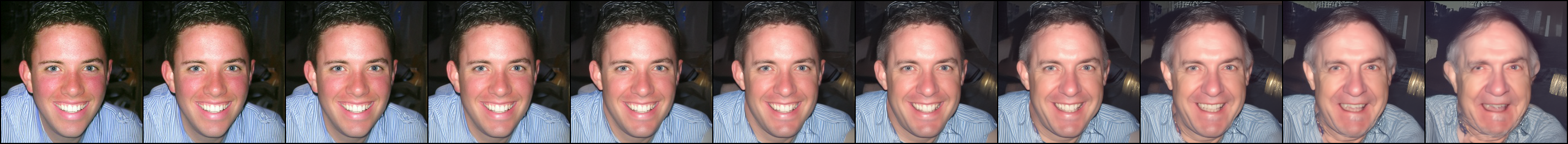}} \\
        \multicolumn{2}{c}{$\Leftarrow$ Young} & & & & \multicolumn{1}{c}{Original} & & & & \multicolumn{2}{c}{Old $\Rightarrow$} \\
        \end{tabularx}
        \caption{Qualitative manipulation results on \textit{skin color} and \textit{age} (protected characteristics) using the proposed ContraCLIP+DiffAE.}
        \label{fig:qualitative_results}
         \vspace*{-0.3cm}
    \end{figure*}

\section{Vision-Language Bias Control (VLBC)}\label{sec:vlbc}
    
    In this section, we present our method for controlling the bias of a given dataset via augmenting images by editing specific protected characteristics (e.g., \textit{skin colour}) using a diffusion-based generative model (DiffAE~\cite{preechakul2021diffusion}) and an augmentation module that learns to generate images driven by prompts in natural language (ContraCLIP~\cite{tzelepis2022contraclip}). An overview of the proposed framework, which we call Vision-Language Bias Control (VLBC), is shown in Fig.~\ref{fig:overview}. Concretely, given a real training dataset $\mathcal{X}_r$ of facial images, annotated for several attribute classes (e.g., CelebA-HQ's~\cite{liu2015faceattributes} attributes, such as \textit{chubby}, \textit{long hair}, etc.), we first calculate, for each class, the number of positive and negative samples with respect to a protected characteristic (e.g., \textit{skin colour}). By doing so, we identify whether the protected characteristic at hand is under-represented in the given dataset. Then, after having identified the bias towards a specific protected characteristic, we may control it (i.e., mitigate or increase it) by i) selecting fake images from a large datasets of synthetic facial images generated by DiffAE~\cite{preechakul2021diffusion}, $\mathcal{X}_s$, that have been pseudo-labelled by a pre-trained network on 41 attributes $f$~\cite{liu2015faceattributes,karkkainenfairface} (\textit{Synthetic Image Selection}) and ii) manipulating accordingly using the proposed \textit{Augmentation Module}. The augmented dataset, $\mathcal{X}_a$, is then used along with the original (biased) dataset $\mathcal{X}_r$ towards training fairer downstream classifiers. We note that we do not merely sample synthetic images from $\mathcal{X}_s$ since we do not possess any control over the attributes of the generated images and there is no guarantee that generated images will be numerically adequate to compensate for the under-represented classes of protected characteristics. This is simply because synthetic images follow the dataset distribution where the generative models have been trained, thus, they still suffer from biases that are present in those datasets. We prove this intuition by reporting the failing of this method on \textit{skin color} (see \textit{Baseline-sampling} Sect.\ref{subsec:mitigating_bias} and Table \ref{table:CelebAHQ_skin_color}).

    \subsection{Synthetic image selection module}\label{subsec:img_selection}

        In this section we present the \textit{Synthetic Image Selection Module} of our framework (see top-left part of Fig.~\ref{fig:overview}). As discussed above, given a real dataset $\mathcal{X}_r$, our goal towards mitigating the bias with respect to a specific protected characteristic is to augment the dataset with images that exemplify that protected characteristic ($\mathcal{X}_a$), so as the resulting dataset ($\mathcal{X}_r+\mathcal{X}_a$) is fairer in a given downstream task.
        
        The aforementioned images that will be manipulated to exemplify the desired protected characteristic are selected from a synthetic dataset $\mathcal{X}_s$ of images generated by the DiffAE~\cite{preechakul2021diffusion}. The images of $\mathcal{X}_s$ are pseudo-labelled by a network $f$ pre-trained on the CelebA's~\cite{liu2015faceattributes} 40 facial attributes (40 binary pseudo-labels) and on FairFace's~\cite{karkkainenfairface} \textit{skin colour} prediction (1 label). We note that while FairFace~\cite{karkkainenfairface} provides predictions with respect to \textit{skin colour} in terms of four groups (i.e., ``white'', ``black'', ``Asian'', and ``Indian''), we use a binary label corresponding to \textit{white}/\textit{black} by removing data from the ``Indian'' group. The resulting annotated synthetic dataset $\mathcal{X}_s$ is given as $\mathcal{X}_s=\{(x_i,s_i,\hat{y}_i)\}_{i=1}^{N_s}$, where $N_s$ denotes the dataset size (in our experiments we set $N_s=120,000$), $x_i$ denotes the $i$-th image and $s_i$ its code in the semantic space of DiffAE~\cite{preechakul2021diffusion}.
        
        Finally, in order to decide on the sort of images required for augmentation (image selection), we calculate the statistics over the real (training) set $\mathcal{X}_r$ by taking into account both protected characteristics and classification attributes. Concretely, we calculate the number of samples that are annotated for the desired classification attribute and that appear as positive or negative with respect to the desired protected characteristic. In the case of \textit{bias mitigation} (towards reducing the bias), we increase the number of images of the minority protected characteristic class, keeping unaltered the downstream attribute labels, by selecting synthetic images $x_i$ for which the pseudo-label $f(x_i)$ is the majority class (e.g., white people see Fig.~\ref{fig:overview}) -- we denote this variant VLBC-. On the contrary, in the case of \textit{bias amplification} (towards increasing the bias), we select to manipulate synthetic images pseudo-labeled with the minority class for the protected characteristic class at hand -- we denote this variant of our method VLBC+. We note that bias mitigation is typically the desired behaviour of the resulting augmented dataset, however our formulation allows for increasing the bias as well, which is designed for research purposes only, towards studying bias in real datasets and downstream tasks. After selecting which images to manipulate towards the desired protected characteristic, we apply the proposed augmentation module, which we describe below.
        
        \begin{table}[t]
            \begin{center}
            \tiny
            \begin{tblr}{@{}c@{}c@{}c@{}c@{}}
                \hline
                \SetCell[c=3]{c} \textbf{Semantic dipoles text-prompts} \\ 
                \textbf{Negative direction ($-$)} & & \textbf{Positive direction ($+$)} \\ \hline
                Young & $\leftrightarrow $ & Old \\
                "An ID photo of a young person." & $\leftrightarrow $ & "An ID photo of an old person." \\ \hline
                White skin color & $\leftrightarrow $ & Black skin color \\
                "A pale skin face." & $\leftrightarrow $ & "A black face." \\ \hline
            \end{tblr}
            \end{center}
            \vspace{-0.4cm}
            \caption{Text prompts used to learn the described manipulations.}
            \label{tab:semantic_dipoles}
            \vspace{-0.4cm}
        \end{table}

    \begin{table}[t]
        \centering
        \tiny
        \begin{tabular}{c@{\hspace{0.3cm}}c@{\hspace{0.3cm}}c@{\hspace{0.3cm}}c@{\hspace{0.3cm}}c@{\hspace{0.3cm}}c@{\hspace{0.3cm}}c}
            \hline
            \multirow{2}{*}{Task}                                       & \multirow{2}{*}{Method}                                      & \multirow{2}{*}{Accuracy $\uparrow$}   & \multirow{2}{*}{f1-score $\uparrow$}      & \multirow{2}{*}{Acc Diff}      & \multicolumn{2}{c}{Fairness}                                                 \\
                                                                        &                                                              &                                        &                                           &                                & $\Delta_A$ $\downarrow$               & $\Delta_M$ $\downarrow$              \\ \hline
            \multirow{7}{*}{\rotatebox[origin=c]{90}{Wearing Necktie}}  & Baseline                                                     & 94.37$\pm$0.11                         & 78.19$\pm$0.24                            & 10.65$\pm$0.31                 & 6.02$\pm$0.27                         & 6.59$\pm$0.54                         \\
                                                                        & Baseline-sampling                                            & 94.29$\pm$0.11                         & 77.46$\pm$0.59                            & 10.97$\pm$0.26                 & 4.59$\pm$1.57                         & 6.19$\pm$2.13                         \\
                                                                        & Weighting \cite{wang2020towards}                             & 94.46$\pm$0.04                         & 75.14$\pm$0.09                            & 11.01$\pm$0.06                 & 5.88$\pm$1.14                         & 9.99$\pm$2.43                         \\
                                                                        & LfF \cite{nam2020learning}                                   & \textbf{94.84$\pm$0.13}                & \textbf{81.11$\pm$0.46}                   & 10.30$\pm$0.30                 & \textbf{4.11$\pm$0.60}                & 6.35$\pm$0.06                         \\
                                                                        & CGL-FairHSIC \cite{jung2022learning}                         & 92.83$\pm$0.00                         & 48.14$\pm$0.00                            & 15.88$\pm$0.00                 &  --                                   & --                                    \\
                                                                        & \textbf{VLBC- (ours)}                                        & 94.69$\pm$0.06                         & 78.48$\pm$0.30                            & \textbf{10.07$\pm$0.13}        & 5.18$\pm$0.35                         & 5.83$\pm$0.75                         \\ 
                                                                        & \textbf{VLBC- $\backslash$f (ours) }                         & 94.65$\pm$0.04                         & 78.43$\pm$0.21                            & 10.18$\pm$0.08                 & 5.21$\pm$0.32                         & \textbf{5.72$\pm$0.47}                    \\\hline
            \multirow{7}{*}{\rotatebox[origin=c]{90}{Chubby}}           & Baseline                                                     & 93.24$\pm$0.04                         & 72.56$\pm$0.88                            & 15.85$\pm$0.65                 & 8.24$\pm$1.42                         & 9.65$\pm$1.92                         \\
                                                                        & Baseline-sampling                                            & 93.16$\pm$0.15                         & 71.6$\pm$0.39                             & \textbf{15.51$\pm$0.48}        & \textbf{3.58$\pm$1.1}                 & \textbf{5.22$\pm$0.53}                         \\
                                                                        & Weighting \cite{wang2020towards}                             & 93.33$\pm$0.09                         & 66.33$\pm$0.72                            & 16.75$\pm$0.21                 & 9.64$\pm$0.57                         & 18.15$\pm$1.38                        \\
                                                                        & LfF \cite{nam2020learning}                                   & 92.60$\pm$0.51                         & \textbf{76.74$\pm$0.74}                   & 15.76$\pm$0.94                 & 18.28$\pm$2.65                        & 21.77$\pm$3.61                        \\ 
                                                                        & CGL-FairHSIC \cite{jung2022learning}                         &  92.49$\pm$0.01                        &  48.93$\pm$0.82                           & 19.41$\pm$0.10                 &  --                                   & --                                    \\
                                                                        & \textbf{VLBC- (ours)}                                        & \textbf{93.44$\pm$0.02}                & 71.96$\pm$0.45                            & 15.69$\pm$0.22                 & 5.39$\pm$0.37                         & 5.79$\pm$0.28                         \\ 
                                                                        & \textbf{VLBC- $\backslash$f (ours)}                          & 93.4$\pm$0.06                          & 71.91$\pm$0.28                            & 15.88$\pm$0.2                  & 4.4$\pm$0.35                          & 5.86$\pm$0.06                         \\  \hline
            \multirow{7}{*}{\rotatebox[origin=c]{90}{Arched Eyebrows}}  & Baseline                                                     & 80.15$\pm$0.3                          & 78.91$\pm$0.34                            & -9.00$\pm$0.20                 & 9.22$\pm$0.12                         & 12.41$\pm$0.36                        \\
                                                                        & Baseline-sampling                                            & 80.18$\pm$0.22                         & 79.15$\pm$0.22                            & -8.25$\pm$0.17                 & 9.73$\pm$0.22                         & 12.75$\pm$0.22                        \\
                                                                        & Weighting \cite{wang2020towards}                             & 79.57$\pm$0.13                         & 78.29$\pm$0.14                            & -9.18$\pm$0.29                 & \textbf{6.25$\pm$0.26}                & \textbf{9.01$\pm$0.45}                \\
                                                                        & LfF \cite{nam2020learning}                                   & 25.27$\pm$0.21                         & 25.06$\pm$0.21                            & 9.36$\pm$0.50                  & --                                    & --                        \\ 
                                                                        & CGL-FairHSIC \cite{jung2022learning}                         & \textbf{84.17$\pm$0.46}                &  \textbf{83.28$\pm$0.46}                  & \textbf{-7.63$\pm$0.48}        & 6.45$\pm$1.38                         & 10.35$\pm$1.28                        \\
                                                                        & \textbf{VLBC- (ours)}                                        & 80.44$\pm$0.06                         & 79.27$\pm$0.04                            & -8.62$\pm$0.19                 & 10.56$\pm$0.20                        & 13.00$\pm$0.53                        \\
                                                                        & \textbf{VLBC- $\backslash$f (ours)}                          & 80.56$\pm$0.14                         & 79.39$\pm$0.14                            & -8.47$\pm$0.28                 & 10.39$\pm$0.04                        & 12.75$\pm$0.55                        \\ \hline
            \multirow{7}{*}{\rotatebox[origin=c]{90}{Double Chin}}      & Baseline                                                     & 94.17$\pm$0.13                         & 74.47$\pm$0.30                            & 15.15$\pm$0.6                  & 18.74$\pm$1.38                        & 27.86$\pm$2.94                        \\
                                                                        & Baseline-sampling                                            & 94.36$\pm$0.23                         & 74.21$\pm$0.58                            & \textbf{13.71$\pm$0.85}        & 10.87$\pm$2.25                        & 15.4$\pm$4.51                         \\
                                                                        & Weighting \cite{wang2020towards}                             & \textbf{94.66$\pm$0.09}                & 69.09$\pm$0.78                            & 14.46$\pm$0.25                 & \textbf{1.69$\pm$0.34}                & \textbf{2.00$\pm$0.24}                 \\
                                                                        & LfF \cite{nam2020learning}                                   & 94.16$\pm$0.08                         & \textbf{78.39$\pm$0.29}                   & 14.27$\pm$0.13                 & 21.94$\pm$1.01                        & 30.54$\pm$1.89                        \\ 
                                                                        & CGL-FairHSIC \cite{jung2022learning}                         & 93.74$\pm$0.00                         &  48.39$\pm$0.00                           &  18.67$\pm$0.00                & --                                    & --                                        \\
                                                                        & \textbf{VLBC- (ours)}                                        & 94.48$\pm$0.04                         & 74.35$\pm$0.23                            & 14.72$\pm$0.21                 & 14.20$\pm$0.07                        & 20.69$\pm$0.18                        \\
                                                                        & \textbf{VLBC- $\backslash$f (ours)}                          & 94.46$\pm$0.04                         & 74.19$\pm$0.33                            & 14.57$\pm$0.15                 & 15.01$\pm$0.25                        & 22.44$\pm$0.48                            \\\hline
        \end{tabular}
        \caption{Results of the classification tasks with \textbf{age} as protected characteristic. We train the model with the original training data (baseline), with the original data plus synthetic (baseline-sampling), and with the proposed VLBC-. We compare with \textit{weighting}~\cite{wang2020towards}, LfF~\cite{nam2020learning}, and CGL-FairHSIC~\cite{jung2022learning}.}
        \label{table:CelebAHQ_age}
         \vspace*{-0.4cm}
    \end{table}
    
    \subsection{Augmentation module (ContraCLIP+DiffAE)}\label{subsec:aug_module}

        In this section, we present the proposed augmentation module that is capable of manipulating facial images with respect to a protected characteristic (e.g., \textit{skin color}) described in natural language. For doing so, we build on the work of Tzelepis et al.~\cite{tzelepis2022contraclip} and we modify ContraCLIP so that it discovers language-driven controllable paths in the semantic space of DiffAE~\cite{preechakul2021diffusion}, instead of that of StyleGAN2~\cite{stylegan2_karras20cvpr}. We briefly discuss both components below.
        
        \paragraph{Diffusion AutoEncoder (DiffAE)~\cite{preechakul2021diffusion}} has been recently proposed to endow a Denoising Diffusion Implicit Model (DDIM)~\cite{song2020denoising} with a meaningful semantic space $\mathcal{S}$ and, thus, the ability of semantic editing.

        \paragraph{ContraCLIP~\cite{tzelepis2022contraclip}} Given a pre-trained GAN, ContraCLIP~\cite{tzelepis2022contraclip} learns non-linear paths in its latent space driven by contrasting semantic dipoles given in natural language. To do so, one needs to define pairs of sentences that convey contrasting meanings and express the limits of the interpretation that are required by the optimized latent paths to encode. Each such pair corresponds to one trainable path in the GAN latent space. Given $K$ semantic dipoles, ContraCLIP first represents them in the CLIP text space and then use them as the centres of $K$ RBF-based warping functions~\cite{tzelepis2022contraclip} $\{\zeta_k\}_{k=1}^{K}$. This gives rise to $K$ vector fields, which provide paths from the one pole/sentence to the other and can be used as the supervisory signal that guides the trainable paths, for any given image and its transformed version along a certain latent path.

        In this work, we introduce our augmentation module (illustrated in the bottom-right part of Fig.~\ref{fig:overview}) by building on ContraCLIP~\cite{tzelepis2022contraclip} and extending it towards learning generative paths in the semantic space $\mathcal{S}$ of DiffAE~\cite{preechakul2021diffusion}. We remark that, in our pipeline, we do not train the DiffAE's~\cite{preechakul2021diffusion} generator $\mathcal{G}$ (that is \textit{pre-trained} on FFHQ~\cite{Karras_2019_CVPR}). Hereafter, we will be referring to our augmentation module as ContraCLIP+DiffAE. We pre-train ContraCLIP+DiffAE to learn paths that refer to protected characteristics (see Tab.~\ref{tab:semantic_dipoles}). Then, given the semantic code $s_i\in\mathcal{S}$ of the $i$-th synthetic image, we can manipulate it by traversing across the $k$-th path (that corresponds to the $k$-th protected characteristic; e.g., \textit{skin colour}) by performing steps given by $s_i^{\prime} = s_i + \epsilon \zeta_k(s_i)$, where $\zeta_k$ is the warping function~\cite{tzelepis2022contraclip} for the $k$-th protected characteristic and $\epsilon$ is scalar determining the length and the direction of the manipulation step.

        It is worth noting that the traversal length in the semantic space $\mathcal{S}$ affects the degree of manipulation. Concretely, in order to enforce diversity during the augmentation phase and guarantee that manipulation is effective (i.e., it changes the characteristic at hand adequately), we define a minimum ($\mathcal{E}_{\min}$) and a maximum ($\mathcal{E}_{\max}$) number of traversal steps, and we randomly (uniformly) sample the number of steps ($\mathcal{E}$) in $[\mathcal{E}_{\min},\mathcal{E}_{\max}]$ at each augmentation. That is, given a synthetic image $x_i\in \mathcal{X}_s$, we manipulate its semantic code by applying $\mathcal{E}$ steps before we arrive at the final augmented image $s_i^a\in \mathcal{S}$. As a result, after manipulating all selected synthetic images $x_i$, we obtain an augmented dataset $\mathcal{X}_a$, that is used along with the real dataset $\mathcal{X}_r$ in order to make it fairer with respect to the downstream task at hand. This is illustrated in the bottom-right part of Fig.~\ref{fig:overview}.

        \begin{table}[t]
            \centering
            \tiny
            \begin{tabular}{c@{\hspace{0.3cm}}c@{\hspace{0.3cm}}c@{\hspace{0.3cm}}c@{\hspace{0.3cm}}c@{\hspace{0.3cm}}c@{\hspace{0.3cm}}c}
                \hline
                \multirow{2}{*}{Task}                                       & \multirow{2}{*}{Method}                                      & \multirow{2}{*}{Accuracy $\uparrow$}   & \multirow{2}{*}{f1-score $\uparrow$}      & \multirow{2}{*}{Acc Diff}      & \multicolumn{2}{c}{Fairness}                                                  \\
                                                                            &                                                              &                                        &                                           &                                & $\Delta_A$ $\downarrow$               & $\Delta_M$ $\downarrow$              \\ \hline
                \multirow{7}{*}{\rotatebox[origin=c]{90}{Straight Hair}}      & Baseline                                                      & 82.52$\pm$0.53                & 71.88$\pm$0.74            & 10.17$\pm$0.81            & 20.16$\pm$3.61                      & 32.85$\pm$6.24              \\
                                                    & Baseline-sampling                                             & 81.96$\pm$0.26                & 72.12$\pm$0.27            & 10.57$\pm$0.34            & 22.17$\pm$2.9                       & 35.38$\pm$5.38              \\
                                                    & Weighting \cite{wang2020towards}                              & 81.88$\pm$0.27                & 71.01$\pm$0.46            & 10.75$\pm$0.76            & 16.53$\pm$3.13                      & 25.61$\pm$5.80              \\
                                                    & LfF \cite{nam2020learning}                                    & 39.50$\pm$3.24                & 39.12$\pm$2.89            & 17.46$\pm$0.61            & --                              & --                            \\
                                                    & CGL-FairHSIC \cite{jung2022learning}    & \textbf{84.51$\pm$0.12}       & \textbf{76.15$\pm$0.61}   & \textbf{9.00$\pm$0.72}    &  15.93$\pm$2.48                     & \textbf{24.65$\pm$5.78}     \\ 
                                                    & \textbf{VLBC- (ours)}                                         & 81.99$\pm$0.08                & 70.78$\pm$0.08            & 9.48$\pm$0.22             & \textbf{15.56$\pm$0.81}             & 25.39$\pm$1.44              \\ 
                                                    & \textbf{VLBC- $\backslash$f (ours)}                              & 82.03$\pm$0.14                & 70.63$\pm$0.3             & 9.44$\pm$0.14             & 15.75$\pm$0.33                      & 25.91$\pm$0.65              \\\hline
                \multirow{7}{*}{\rotatebox[origin=c]{90}{Young}}              & Baseline                                                      & 85.25$\pm$0.09                & 79.43$\pm$0.26            & -9.04$\pm$0.09            & 8.96$\pm$0.30                       & 10.40$\pm$1.20              \\
                                                    & Baseline-sampling                                             & 85.33$\pm$0.36                & 79.8$\pm$0.41             & -10.95$\pm$0.8            & 10.2$\pm$0.74                       & 11.85$\pm$0.5               \\
                                                    & Weighting \cite{wang2020towards}                              & 85.1$\pm$0.09                 & 78.87$\pm$0.23            & -7.23$\pm$1.86            & 8.15$\pm$2.66                       & 11.96$\pm$4.44              \\
                                                    & LfF \cite{nam2020learning}                                    & 21.18$\pm$0.54                & 21.16$\pm$0.56            & 5.51$\pm$1.27             & --                               & --                               \\
                                                    & CGL-FairHSIC \cite{jung2022learning}    & \textbf{87.93$\pm$0.28}       & \textbf{82.25$\pm$0.82}   & -7.57$\pm$1.26            &  7.27$\pm$1.65                      & 10.03$\pm$2.45              \\ 
                                                    & \textbf{VLBC- (ours)}                                         & 85.55$\pm$0.04                & 79.10$\pm$0.27            & \textbf{-6.27$\pm$0.56}   & \textbf{5.32$\pm$0.55}              & \textbf{5.46$\pm$0.59}      \\
                                                    & \textbf{VLBC- $\backslash$f (ours)}                              & 85.5$\pm$0.09                 & 79.27$\pm$0.15            & -7.38$\pm$0.48            & 5.68$\pm$0.29                       & 7.52$\pm$0.75               \\\hline
                \multirow{7}{*}{\rotatebox[origin=c]{90}{Wearing Necktie}}    & Baseline                                                      & 94.35$\pm$0.11                & 77.49$\pm$0.20            & -8.10$\pm$0.63            & 11.00$\pm$1.93                      & 17.13$\pm$3.67          \\
                                                    & Baseline-sampling                                             & 94.4$\pm$0.03                 & 78.17$\pm$0.31            & -7.86$\pm$0.6             & 11.43$\pm$0.94                      & 18.25$\pm$1.82          \\
                                                    & Weighting \cite{wang2020towards}                              & 94.33$\pm$0.12                & 76.74$\pm$0.33            & -5.48$\pm$0.27            & 10.66$\pm$1.15                      & 19.89$\pm$2.30           \\
                                                    & LfF \cite{nam2020learning}                                    & \textbf{94.86$\pm$0.07}       & \textbf{81.27$\pm$0.06}   & -7.49$\pm$0.47            & 15.01$\pm$1.85                      & 26.28$\pm$4.06          \\
                                                    & CGL-FairHSIC \cite{jung2022learning}    & 93.65$\pm$0.78                & 61.18$\pm$10.09           & \textbf{-3.88$\pm$0.41}   &  \textbf{0.32$\pm$0.22}             & \textbf{0.5$\pm$0.28}   \\ 
                                                    & \textbf{VLBC- (ours)}                                         & 94.45$\pm$0.04                & 76.81$\pm$0.32            & -6.38$\pm$0.11            & 3.67$\pm$0.84                       & 4.46$\pm$0.01           \\ 
                                                    & \textbf{VLBC- $\backslash$f (ours)}                              & 94.46$\pm$0.04                & 76.77$\pm$0.19            & -6.29$\pm$0.11            & 3.49$\pm$0.5                        & 4.37$\pm$0.06           \\ \hline
                \multirow{7}{*}{\rotatebox[origin=c]{90}{Big Lips}}           & Baseline                                                      & 65.62$\pm$0.54                & 57.62$\pm$0.12            & 6.32$\pm$1.47             & 45.17$\pm$2.41                      & 54.23$\pm$0.82          \\
                                                    & Baseline-sampling                                             & \textbf{65.78$\pm$0.67}       & \textbf{58.5$\pm$0.87}    & 8.94$\pm$0.96             & 51.82$\pm$1.21                      & 58.33$\pm$0.54          \\
                                                    & Weighting \cite{wang2020towards}                              & 63.32$\pm$0.13                & 58.04$\pm$0.71            & -0.84$\pm$3.48            & \textbf{23.03$\pm$2.35}             & \textbf{28.48$\pm$1.60}  \\
                                                    & LfF \cite{nam2020learning}                                    & 35.45$\pm$0.22                & 34.58$\pm$0.41            & -4.97$\pm$0.57            & --                               & --                       \\
                                                    & CGL-FairHSIC \cite{jung2022learning}    & 65.72$\pm$1.04                & 58.29$\pm$0.93            & \textbf{-0.71$\pm$2.48}   &  35.09$\pm$4.34                     & 40.79$\pm$4.45          \\ 
                                                    & \textbf{VLBC- (ours)}                                         & 65.69$\pm$0.04                & 57.67$\pm$0.14            & 5.77$\pm$0.63             & 43.74$\pm$0.42                      & 53.05$\pm$0.77          \\ 
                                                    & \textbf{VLBC- $\backslash$f (ours)}                              & 65.66$\pm$0.07                & 57.91$\pm$0.33            & 5.02$\pm$0.36             & 41.04$\pm$0.38                      & 50.38$\pm$0.17          \\ \hline
                \multirow{7}{*}{\rotatebox[origin=c]{90}{Big Nose}}           & Baseline                                                      & 78.82$\pm$0.07                & 74.74$\pm$0.21            & 3.53$\pm$0.35             & 29.72$\pm$1.96                      & 31.88$\pm$2.59          \\
                                                    & Baseline-sampling                                             & \textbf{79.0$\pm$0.17}        & \textbf{74.88$\pm$0.35}   & 6.9$\pm$0.52              & 35.45$\pm$0.63                      & 36.71$\pm$0.46          \\
                                                    & Weighting \cite{wang2020towards}                              & 77.77$\pm$0.28                & 73.62$\pm$0.27            & \textbf{1.29$\pm$0.89}    & \textbf{18.71$\pm$0.69}             & \textbf{23.12$\pm$0.77} \\ 
                                                    & LfF \cite{nam2020learning}                                    & 22.92$\pm$0.52                & 22.68$\pm$0.59            & 5.19$\pm$1.26             & --                              & --                        \\
                                                    & CGL-FairHSIC \cite{jung2022learning}    & 78.59$\pm$0.22                & 74.36$\pm$0.41            & 7.73$\pm$0.3              &  36.51$\pm$1.36                     & 37.92$\pm$1.03          \\ 
                                                    & \textbf{VLBC- (ours)}                                         & 78.26$\pm$0.06                & 74.23$\pm$0.12            & 2.40$\pm$0.33             & 22.01$\pm$0.60                      & 25.04$\pm$0.88          \\ 
                                                    & \textbf{VLBC- $\backslash$f (ours)}                              & 78.37$\pm$0.1                 & 74.32$\pm$0.03            & 2.48$\pm$0.18             & 23.41$\pm$0.49                      & 25.78$\pm$0.21          \\ \hline
            \end{tabular}
            \caption{Results of the classification tasks with \textbf{skin color} as protected characteristic. We train the model with the original training data (baseline), with the original data plus synthetic (baseline-sampling), and the proposed VLBC-. We compare with \textit{weighting}~\cite{wang2020towards}, LfF~\cite{nam2020learning}, and CGL-FairHSIC~\cite{jung2022learning}.}
            \label{table:CelebAHQ_skin_color}
            \vspace*{-0.4cm}
        \end{table}

    \section{Experiments}\label{sec:experiments}
    
    In this section, we present the experimental evaluation of the proposed framework for controlling the bias in facial datasets with respect to the protected characteristics \textit{skin color} and \textit{age}, towards the downstream task of binary attribute classification. We note that we train the classification models only on the classification attributes, not the protected characteristic. We provide qualitative and quantitative results on both mitigating (VLBC-, Sect.~\ref{subsec:mitigating_bias}) and increasing (VLBC+, Sect.~\ref{subsec:increase_bias}) the bias, and we compare with the state-of-the-art (SOTA) works of~\cite{wang2020towards,nam2020learning,jung2022learning}. We first train a baseline model on the original training set, quantifying its initial bias without applying any fairness-related technique. Then we investigate how the same model behaves when fine-tuned on the augmented dataset created to mitigate or increase the bias. Moreover, while mitigating the bias, we introduce a second baseline, referred to as \textit{baseline-sampling}, which injects synthetic images of the desired protected characteristic (e.g., black people) in the training set without applying the proposed augmentations. This baseline provides a simple yet effective way of determining whether and when our augmentation scheme is necessary, providing additional insight into the usefulness of synthetic images. As discussed in Sect.~\ref{sec:vlbc}, we draw our intuition from the fact that generative model are inherently biased as well, since they depend on training on biased datasets. Finally, in Sect.~\ref{subsec:ablation}, we present our ablation study.
    
    \vspace{-0.4cm}
    \paragraph{Implementation details} We evaluate our method using \textit{MobileNetV2}~\cite{DBLP:journals/corr/abs-1801-04381,tzelepis2023falco} and we train our models on one Nvidia RTX A6000 with SGD and Focal Loss~\cite{lin2017focal}. Learning rate is $10^{-3}$ when training from scratch (starting with ImageNet~\cite{deng2009imagenet} weights) and $10^{-4}$ when fine-tuning. The baselines are trained for $100$ epochs while fine-tuning (VLBC) is performed for $50$ epochs for both \textit{age} and \textit{skin color}. We average the results over three runs and report mean/std. 
    
    \vspace{-0.4cm}
    \paragraph{Datasets} We evaluate the proposed framework on (i) \textbf{CelebA-HQ}~\cite{liu2015faceattributes}, a diverse dataset in terms of \textit{skin colour}, \textit{gender}, and \textit{age}, which contains $30,000$ images annotated with 40 attributes, and (ii) \textbf{UTKFace}~\cite{zhifei2017cvpr}, an in-the-wild dataset consisting of $18,038$ face images annotated with respect to \textit{gender}, \textit{age} (spanning 116 years), and \textit{skin colour}.
    
    \vspace{-0.4cm}
    \paragraph{Evaluation metrics} Under the binary classification setting, a natural way for describing fairness is to have a model performing \textit{equally} regardless of the protected characteristic. For instance, predicting \textit{big lips} should perform independently of characteristics such as \textit{age} or \textit{skin color}. Following this intuition, we calculate the accuracy of the downstream task conditioned on the protected characteristic~\cite{xu2018fairgan}. E.g., in the case of the protected characteristic of \textit{age}, we split the attribute classification accuracy into ``young'' and ``old''. We then calculate the difference between the two accuracies to capture the model's fairness. Ideally, a model will exhibit equal behavior on a zero-valued difference in accuracy. We also note that the sign of the difference in accuracy is indicative of the ``direction'' of the bias -- i.e., a negative difference value would indicate a bias towards elder people, and vice versa for a positive value. We report the overall accuracy, the f1-score, and the difference in accuracy (Acc Diff). Additionally, we calculate the mean $\left(\Delta_A\right)$ and max $\left(\Delta_M\right)$ disparity of opportunity similarly to~\cite{jung2022learning}.    

    \begin{table}
        \centering
        \tiny
        \begin{tabular}{ccccc}
            \multicolumn{5}{c}{\textbf{Age}}                                                                                      \\ \hline
            Task                                                & Method                                                        & Accuracy $\uparrow$               & f1-score $\uparrow$                   & Acc Diff                  \\ \hline
            \multirow{7}{*}{\rotatebox[origin=c]{90}{Gender}}   & Baseline                                                      & 82.87$\pm$0.34                    & 82.86$\pm$0.34                        & 4.8$\pm$1.72             \\
                                                                & Baseline-sampling                                             & 83.85$\pm$0.43                    & 83.84$\pm$0.43                        & 4.83$\pm$0.71                \\                             
                                                                & Weighting \cite{wang2020towards}                              & 83.78$\pm$0.12                    & 83.77$\pm$0.13                        & 6.17$\pm$0.68            \\
                                                                & LfF \cite{nam2020learning}                                    & 44.72$\pm$3.19                    & 43.61$\pm$3.43                        & -2.9$\pm$0.93                      \\
                                                                & CGL-FairHSIC \cite{jung2022learning}                          & \textbf{91.9$\pm$1.0}             & \textbf{91.9$\pm$1.0}                 & \textbf{3.13$\pm$0.83}     \\ \cline{2-5}
                                                                & \textbf{VLBC- (ours)}                                         & 82.73$\pm$0.14                    & 82.73$\pm$0.14                        & 4.67$\pm$0.45            \\ 
                                                                & \textbf{VLBC- $\backslash$f (ours)}                              & 82.7$\pm$0.11                     & 82.69$\pm$0.11                        & 3.93$\pm$0.54             \\\hline
            \multicolumn{5}{c}{\textbf{Skin Color}}                                                                                                                                                                                     \\ \hline
            Task                                                & Method                                                        & Accuracy $\uparrow$               & f1-score $\uparrow$                   & Acc Diff                  \\ \hline
            \multirow{7}{*}{\rotatebox[origin=c]{90}{Gender}}   & Baseline                                                      & 83.57$\pm$0.48                    & 83.56$\pm$0.48                        & 2.6$\pm$1.1              \\
                                                                & Baseline-sampling                                             & 84.53$\pm$0.13                    & 84.53$\pm$0.13                        & 3.4$\pm$0.24                  \\
                                                                & Weighting \cite{wang2020towards}                              & 83.3$\pm$0.28                     & 83.3$\pm$0.28                         & 4.2$\pm$0.75             \\
                                                                & LfF \cite{nam2020learning}                                    & 45.47$\pm$0.82                    & 44.21$\pm$1.21                        & 0.0$\pm$1.07                         \\
                                                                & CGL-FairHSIC \cite{jung2022learning}                          & \textbf{91.28$\pm$0.26}           & \textbf{91.28$\pm$0.26}               & 3.17$\pm$1.09                \\ \cline{2-5}
                                                                & \textbf{VLBC- (ours)}                                         & 83.08$\pm$0.18                    & 83.06$\pm$0.18                        & \textbf{0.03$\pm$0.31}   \\ 
                                                                & \textbf{VLBC- $\backslash$f (ours)}                           & 83.0$\pm$0.25                     & 82.98$\pm$0.24                        & 0.47$\pm$0.56                 \\\hline
            \multirow{7}{*}{\rotatebox[origin=c]{90}{Age}}      & Baseline                                                      & 76.25$\pm$0.54                    & 76.05$\pm$0.57                        & -12.1$\pm$1.06           \\
                                                                & Baseline-sampling                                             & 76.82$\pm$0.51                    & 76.65$\pm$0.53                        & -12.83$\pm$0.12              \\
                                                                & Weighting \cite{wang2020towards}                              & 76.17$\pm$0.09                    & 76.03$\pm$0.12                        & \textbf{-9.93$\pm$0.09}  \\
                                                                & LfF \cite{nam2020learning}                                    & 32.52$\pm$0.73                    & 30.81$\pm$0.6                         & 10.17$\pm$1.76                    \\
                                                                & CGL-FairHSIC \cite{jung2022learning}                          & \textbf{80.55$\pm$1.67}           & \textbf{80.32$\pm$1.86}               & -11.23$\pm$0.4              \\ \cline{2-5}
                                                                & \textbf{VLBC- (ours)}                                         & 76.52$\pm$0.06                    & 76.31$\pm$0.05                        & -11.7$\pm$0.45           \\ 
                                                                & \textbf{VLBC- $\backslash$f (ours)}                           & 76.35$\pm$0.43                    & 76.15$\pm$0.44                        & -11.97$\pm$0.46               \\\hline
        \end{tabular}
        \caption{Results on the UTKFace dataset (\textit{age} and \textit{skin color}).}
        \vspace*{-0.4cm}
        \label{table:utkface_results}
    \end{table}
    \subsection{Bias mitigation with VLBC-}\label{subsec:mitigating_bias}

        In this section, we show how the proposed framework for mitigating the bias (VLBC-) improves the fairness of a given dataset, assesed on a subset of the attribute classification tasks of CelebA-HQ~\cite{liu2015faceattributes}. We chose this based on the bias of a baseline model trained on all the CelebA-HQ attributes. That is, we ranked the tasks based on the difference in accuracy, and we chose the ones with higher values. Specifically, we decided to evaluate our method on the attribute classification tasks of \textit{wearing necktie}, \textit{chubby}, \textit{arched eyebrows}, and \textit{double chin} having \textit{age} as the protected characteristic, and on \textit{straight hair}, \textit{young}, \textit{wearing necktie}, \textit{big lips}, and \textit{big nose} having skin color as the protected characteristic. We employed our augmentation module (ContraCLIP+DiffAE) to balance the training set statistics with respect to the protected characteristics. For example, given big nose and skin color as pair attribute-protected characteristics, we sample images from $\mathcal{X}_s$ of white people (majority class) with and without \textit{big nose} editing them into black people (minority class) balancing the training set. 
        
        We show the results in Tab.~\ref{table:CelebAHQ_age} and \ref{table:CelebAHQ_skin_color}, where we compare the baselines with the following SOTA works: Wang et al.~\cite{wang2020towards} (weighting), Learning from Failure~\cite{nam2020learning} (LfF) and Fairness with the Partially annotated Group labels~\cite{jung2022learning} (CGL-FairHSIC). CGL-FairHSIC proposes to improve fairness by incorporating a partially annotated dataset, thus we apply it to the synthetic dataset. Please note that we denote with a ``-'' the $\left(\Delta_A\right)$ and $\left(\Delta_M\right )$ metrics when the model collapses (e.g, f1-score of LfF~\cite{nam2020learning} and CGL-FairHSIC~\cite{jung2022learning} in some cases). The results show how the proposed framework (VLBC-) is always capable of mitigating bias with respect to the baseline model on all attributes and metrics exhibiting \textit{consistency} over multiple settings. The comparison with SOTA methods highlights how other works are robust in some settings but fail in others. Specifically, Wang et al.~\cite{wang2020towards} (weighting) deteriorates the model's fairness in \textit{wearing necktie}, \textit{arched eyebrows}, and \textit{chubby} attributes when \textit{age} is the protected characteristic. LfF~\cite{nam2020learning} is performing poorly, achieving similar or worse bias than the baseline model trained only on the original data and showing a clear drop in accuracy and f1-score with \textit{skin color} as the protected characteristic. CGL-FairHSIC~\cite{jung2022learning} is performing well when \textit{skin color} is the protected characteristic, but fails on \textit{age}. We argue that weighting~\cite{wang2020towards} fails since it possibly uses for training multiple instances of the same image from the minority class. 
        LfF~\cite{nam2020learning} learns the de-biased model by presenting hard samples coming from a second model specifically trained for increasing the bias potentially challenging the model leading to worse performance (see Table \ref{table:CelebAHQ_skin_color}). Finally, CGL-FairHSIC~\cite{jung2022learning} fails because the fairness loss makes the model converge towards negative predictions degrading in this way the performance.
        
        Moreover, as discussed in Sect.~\ref{sec:vlbc}, the ``baseline-sampling'' approach mitigates the bias only when enough images are available for balancing the training set -- this is clear in the case of \textit{age} (see Tab.~\ref{table:CelebAHQ_age}). In this setting, the diffusion model has a low bias against \textit{age}, thus it generates enough images for both young and old people to balance the original training set. By contrast, the bias mitigation performance worsens when testing the same approach on \textit{skin color} (see Tab.~\ref{table:CelebAHQ_skin_color}), since not enough images of black people are generated by the generative model. These empirical results confirm our intuition that augmentation is required when samples from the minority class are not enough.
        
        In terms of overall performance, our method is able to mitigate the bias while preserving or improving the accuracy and f1-score of the model, demonstrating its effectiveness. In Tab.~\ref{table:utkface_results}, we show the results of bias mitigation applied on the UTKFace~\cite{zhifei2017cvpr} dataset with \textit{age} and \textit{skin color} as protected characteristics. We note that our method is \textit{consistent} also in this dataset mitigating the bias while preserving the performance. Weighting~\cite{wang2020towards} increases the bias on two attributes out of three and LfF~\cite{nam2020learning} does not preserve the overall performance. CGL-FairHSIC~\cite{jung2022learning} worsens the bias performance on \textit{gender} as attribute and \textit{skin color} as protected characteristic, while showing satisfactory performance on the other two settings. Finally, the ``baseline-sampling'' approach maintains the existing bias on \textit{age}, while worsening it on \textit{skin color} similarly to what was discussed above. As a final remark, we note that filtering out the images with incorrect protected characteristic augmentation has a low impact on the results (VLBC- $\backslash$f). This is due to a low error rate during the augmentations. 
    
    \subsection{Increasing the bias with VLBC+}\label{subsec:increase_bias}

        \begin{table}[t]
            \vspace*{-0.2cm}
            \centering
            \tiny
            \begin{tabular}{cccc}
                \multicolumn{4}{c}{\textbf{Age}}                                                                                      \\ \hline
                Task               & Method                 & Accuracy $\uparrow$                      & Acc Diff           \\ \hline
                \multirow{2}{*}{\rotatebox[origin=c]{0}{\shortstack{Wearing \\Necktie}}} & Baseline            &   94.37$\pm$0.11	& 10.65$\pm$0.31            \\
                & \textbf{VLBC+ (ours)} & 94.35$\pm$0.06	& \textbf{11.15$\pm$0.01} \\ \hline
                \multirow{2}{*}{\rotatebox[origin=c]{0}{Chubby}} & Baseline               & 93.24$\pm$0.04	& 15.85$\pm$0.65             \\
                & \textbf{VLBC+ (ours)} & 93.01$\pm$0.05	& \textbf{17.58$\pm$0.09} \\ \hline
                \multirow{2}{*}{\rotatebox[origin=c]{0}{\shortstack{Arched \\Eyebrows}}} & Baseline               & 80.15$\pm$0.3	& \textbf{-9.0$\pm$0.2}             \\
                & \textbf{VLBC+ (ours)} & 80.0$\pm$0.11	& -8.8$\pm$0.18 \\ \hline        
                \multirow{2}{*}{\rotatebox[origin=c]{0}{\shortstack{Double \\Chin}}} & Baseline               & 94.17$\pm$0.13	& \textbf{15.15$\pm$0.6}             \\
                & \textbf{VLBC+ (ours)} & 94.44$\pm$0.05	& 13.73$\pm$0.13 \\ \hline
                \multicolumn{4}{c}{\textbf{Skin Color}}                                                                              \\ \hline
                Task               & Method                 & Accuracy $\uparrow$                        & Acc Diff            \\ \hline
                \multirow{2}{*}{\rotatebox[origin=c]{0}{\shortstack{Big \\Lips}}} & Baseline               & 65.62$\pm$0.54	& 6.32$\pm$1.47             \\
                & \textbf{VLBC+ (ours)} & 65.13$\pm$0.05	& \textbf{11.66$\pm$0.52} \\ \hline
                \multirow{2}{*}{\rotatebox[origin=c]{0}{\shortstack{Big \\Nose}}} & Baseline               &                    78.82$\pm$0.07	& 3.53$\pm$0.35             \\
                & \textbf{VLBC+ (ours)} & 78.68$\pm$0.14	& \textbf{6.57$\pm$0.42} \\ \hline
                \multirow{2}{*}{\rotatebox[origin=c]{0}{\shortstack{Straight \\Hair}}} & Baseline               & 82.52$\pm$0.53	& \textbf{10.17$\pm$0.81}            \\
                & \textbf{VLBC+ (ours)} & 82.92$\pm$0.01	& 9.63$\pm$0.01 \\ \hline
                \multirow{2}{*}{\rotatebox[origin=c]{0}{Young}} & Baseline               &                    85.25$\pm$0.09	& \textbf{-9.04$\pm$0.09}              \\
                & \textbf{VLBC+ (ours)} & 85.52$\pm$0.09	& -7.59$\pm$0.38 \\ \hline
                \multirow{2}{*}{\rotatebox[origin=c]{0}{\shortstack{Wearing \\Necktie}}} & Baseline               &                  94.35$\pm$0.11	& \textbf{-8.1$\pm$0.63}               \\
                & \textbf{VLBC+ (ours)} & 94.3$\pm$0.03	& -6.98$\pm$0.34 \\ \hline
            \end{tabular}
            \caption{Results of VLBC employed to increase bias (VLBC+).}
            \label{table:increase_bias}
            \vspace*{-0.4cm}
        \end{table}

        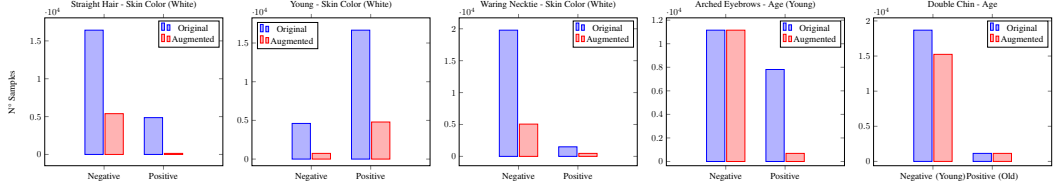
\begin{figure*}[t]
        \centering
            \resizebox{0.8\linewidth}{!}{
                \begin{tabular}{@{}ccccc@{}}
                    \begin{tikzpicture}
                        \begin{axis}[
                            title = {Straight Hair - Skin Color (White)},
                            x tick label style  = {text width=2.5cm,align=center},
                            symbolic x coords={Negative, Positive},
                            ylabel=N° Samples,
                            enlarge x limits=1,
                            ybar,
                            xtick=data,
                            bar width=20pt,
                        ]
                            \addplot coordinates {
                                (Negative, 16414)
                                (Positive, 4863)
                            };
                            \addplot coordinates {
                                (Negative, 5387) 
                                (Positive, 140)
                            };
                            \legend{Original, Augmented}
                        \end{axis}
                    \end{tikzpicture} &    
                    \begin{tikzpicture}
                        \begin{axis}[
                            title = {Young - Skin Color (White)},
                            x tick label style  = {text width=2.5cm,align=center},
                            symbolic x coords={Negative, Positive},
                            enlarge x limits=1,
                            legend pos=north west,
                            ybar,
                            xtick=data,
                            bar width=20pt,
                        ]
                            \addplot coordinates {
                                (Negative, 4607)
                                (Positive, 16660)
                            };
                            \addplot coordinates {
                                (Negative, 736) 
                                (Positive, 4791)
                            };
                            \legend{Original, Augmented}
                        \end{axis}
                    \end{tikzpicture} &
                    \begin{tikzpicture}
                        \begin{axis}[
                            title = {Waring Necktie - Skin Color (White)},
                            x tick label style  = {text width=2.5cm,align=center},
                            symbolic x coords={Negative, Positive},
                            enlarge x limits=1,
                            ybar,
                            xtick=data,
                            bar width=20pt,
                        ]
                            \addplot coordinates {
                                (Negative, 19779)
                                (Positive, 1498)
                            };
                            \addplot coordinates {
                                (Negative, 5058) 
                                (Positive, 469)
                            };
                            \legend{Original, Augmented}
                        \end{axis}
                    \end{tikzpicture} &
                    \begin{tikzpicture}
                        \begin{axis}[
                            title = {Arched Eyebrows - Age (Young)},
                            x tick label style  = {text width=2.5cm,align=center},
                            symbolic x coords={Negative, Positive},
                            enlarge x limits=1,
                            ybar,
                            xtick=data,
                            bar width=20pt,
                        ]
                            \addplot coordinates {
                                (Negative, 11151)
                                (Positive, 7812)
                            };
                            \addplot coordinates {
                                (Negative, 11151) 
                                (Positive, 688)
                            };
                            \legend{Original, Augmented}
                        \end{axis}
                    \end{tikzpicture} &
                    \begin{tikzpicture}
                        \begin{axis}[
                            title = {Double Chin - Age}, 
                            x tick label style  = {text width=3cm,align=center},
                            symbolic x coords={Negative (Young), Positive (Old)},
                            enlarge x limits=1,
                            ybar,
                            xtick=data,
                            bar width=20pt,
                            ]
                            \addplot coordinates {
                                (Negative (Young), 18684)
                                (Positive (Old), 1143)
                            };
                            \addplot coordinates {
                                (Negative (Young), 15242) 
                                (Positive (Old), 1143)
                            };
                            \legend{Original, Augmented}
                        \end{axis}
                    \end{tikzpicture}
                \end{tabular}
            }
            \caption{Distribution of the number of samples for the failure cases of Tab.~\ref{table:increase_bias}.}
            \label{fig:n_samples_increase_bias}
            \vspace*{-0.4cm}
        \end{figure*}

        \textbf{Remark:} \textit{The experiments done in this section are for scientific purposes only and we discourage increasing the bias against specific ethical groups or sensitive attributes.}

        As discussed in previous sections, the proposed framework is also capable of increasing the bias towards a specific attribute/protected characteristic, due to its versatile augmentation module (ContraCLIP+DiffAE). We denote this variant of our method as VLBC+. A useful scenario might be that of increasing the downstream task accuracy on particular attributes or the augmentation of a given dataset by generating faces with specific attributes. However, we stress that increasing bias towards specific attributes must be carefully considered and justified to avoid discriminative and unfair practices. For the sake of coherence, we report here the results of this setting applied to CelebA-HQ~\cite{liu2015faceattributes} on the same attributes and protected characteristics discussed above (Sect.~\ref{subsec:mitigating_bias}). In this scenario, we augment the images to unbalance, even more, the majority class (e.g., white people). We aim at doubling the majority class; that is, to give enough statistical evidence to the model during training. As we can see in Tab.~\ref{table:increase_bias}, our method (VLBC+) is able to increase the bias on four attributes out of nine. We argue that, similarly to the \textit{sampling-baseline} approach, VLBC+ fails when not enough images are augmented, thus not doubling the majority class. To further investigate this issue, we report, in Fig.~\ref{fig:n_samples_increase_bias}, the number of augmented samples compared to the original training set on the attributes where this method is struggling. Since we are augmenting the majority class (e.g., ``white'' skin colour), we report the number of samples conditioned on the majority protected characteristic class showing the positive or negative number of samples for the downstream classification attribute. For example, the leftmost chart shows the number of samples having (positive) or not having (negative) ``straight hair'' appearing as ``white people'' (majority class). As we can see, we cannot generate enough images to actually double the class, since the synthetic dataset $\mathcal{X}_s$ is itself biased and does not capture rare combinations (\textit{attribute-protected characteristics}) to augment. This is due to the inherent bias of the generative model -- that is, we may not be able to generate enough black people to augment towards white in order to influence the bias. When this does not happen, our method does increase the bias, as expected in this setting.

        \begin{figure}[t]
          \setlength\tabcolsep{0pt}
          \adjustboxset{width=\linewidth,valign=c}
          \tiny
          \centering
          \begin{tabularx}{1\linewidth}{@{}
              l
              X 
              X
              X
            @{}}
            \rotatebox[origin=c]{90}{\textbf{Skin Color}}
            & \includegraphics{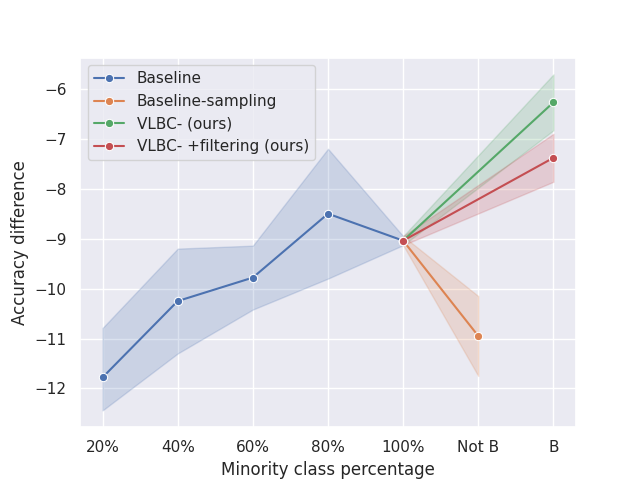}
            & \includegraphics{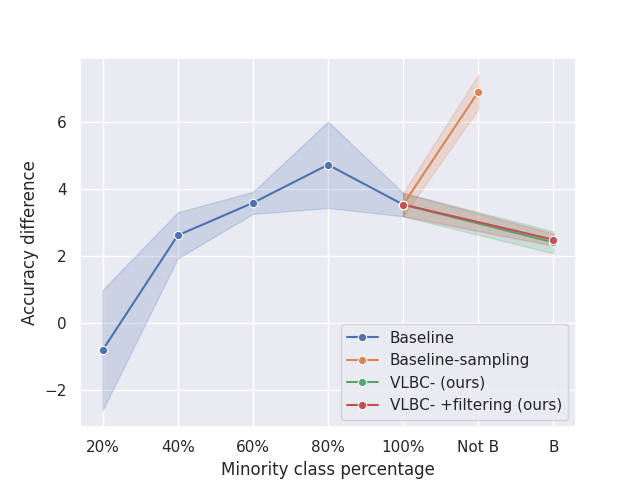}
            & \includegraphics{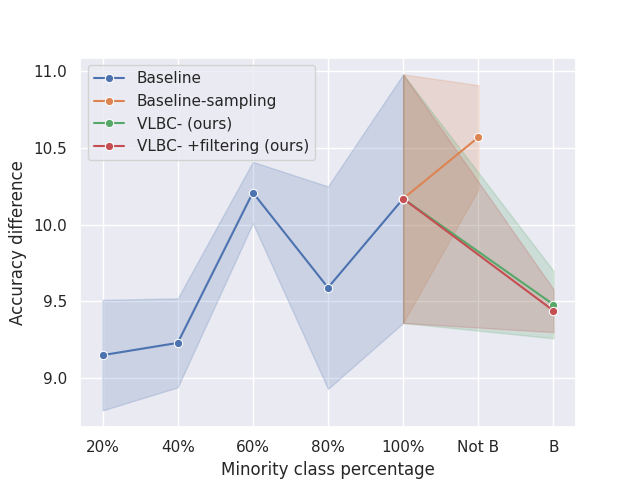} \\
            & \multicolumn{1}{c}{\textbf{Young}}
            & \multicolumn{1}{c}{\textbf{Big Nose}}
            & \multicolumn{1}{c}{\textbf{Straight Hair}} \\
            \rotatebox[origin=c]{90}{\textbf{Age}}
            & \includegraphics{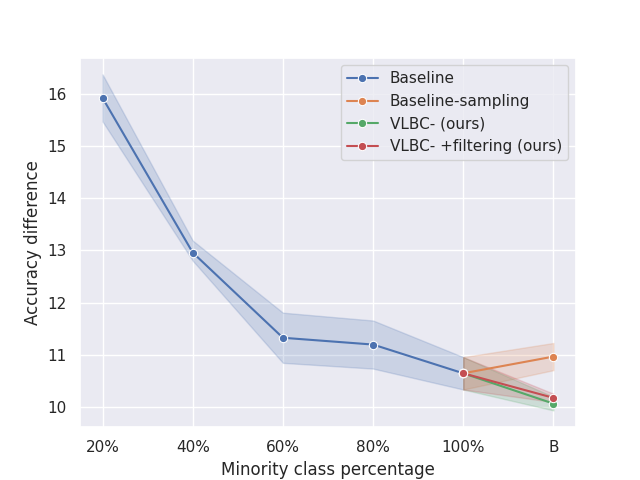}
            & \includegraphics{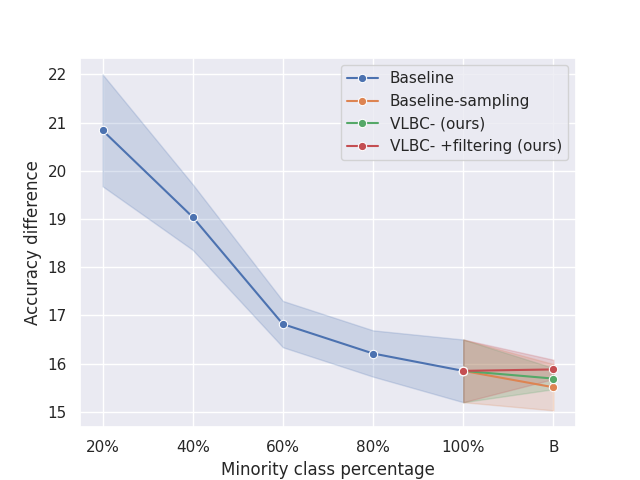}
            & \includegraphics{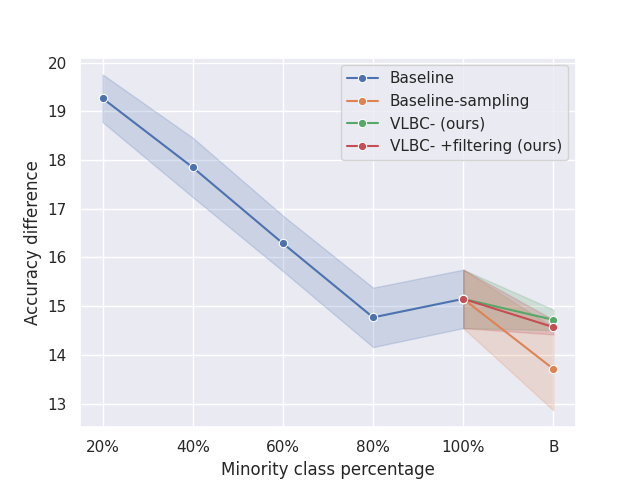} \\
            & \multicolumn{1}{c}{\textbf{Wearing Necktie}}
            & \multicolumn{1}{c}{\textbf{Chubby}}
            & \multicolumn{1}{c}{\textbf{Double Chin}} \\
          \end{tabularx}
          \caption{Accuracy difference for three downstream attributes with \textit{skin color} and \textit{age} as protected characteristics. \textit{B} and \textit{Not B} denote balanced or not balanced data, respectively. Clearly, sampling fails when not enough images are available for sampling.}
          \label{fig:bias_control}
        \end{figure}
    
    \subsection{Ablation study}\label{subsec:ablation}

        We present our ablation on the number of samples of the minority-protected characteristic to show how the bias can be further controlled. We sample $20\%$, $40\%$, $60\%$, $80\%$, $100\%$ (whole training set) of the minority class and, finally, we plug our augmented data to go beyond the $100\%$ balancing the dataset (as in Sect.~\ref{subsec:mitigating_bias}). Fig.~\ref{fig:bias_control} shows this ablation on \textit{Baseline}, \textit{Baseline-sampling}, and \textit{VLBC-} with and without \textit{filtering}, considering three downstream tasks with \textit{skin color} and \textit{age} as the protected characteristics. Fig.~\ref{fig:bias_control} shows a trend of the difference in accuracy proving that adding data to the training set controls the behavior of the model with respect to the protected characteristic. Moreover, this experiment allows us to study whether our method is able to invert the original trend (increasing bias). The graphs report how our method is consistent in mitigating the bias while the sampling method fails on \textit{skin color} due to the balancing issue discussed earlier (please note that on ``Young'' the difference in accuracy is negative, thus we want it to increase towards $0$). We note that, occasionally, training in low data regime (e.g., $20\%$) leads to better fairness (e.g., \textit{straight hair}, \textit{big nose}), but only by sacrificing the accuracy.
    
    \subsection{Qualitative results}\label{subsec:qualitative_results}            

        Given a pre-trained generative path, our augmentation module can control the manipulation strength by means of traversal length. In Fig.~\ref{fig:qualitative_results}, we show the traversals for the two studied protected characteristics (\textit{skin color} and \textit{age}). Clearly, the longer the traversal length, the stronger the manipulation. This demonstrates the effectiveness of the proposed augmentation module (ContraCLIP+DiffAE) in manipulating facial images toward the desired protected characteristics, while at the same time the overall image quality and other facial attributes are preserved. 

\section{Limitations}
    
    Our work exhibits potential limitations due to the assumptions that: (i) the learnt latent paths convey the desired manipulation while preserving the downstream attribute (disentanglement), and (ii) a good pseudo-labelling module is employed. For (i), we attempt to impose the orthogonality of the paths by employing a contrastive loss which improves their disentanglement, while for (ii) we experimentally show (Sect.~\ref{sec:experiments}) that accuracy remains stable across different settings, suggesting that the proposed framework exhibits robustness even with a simple pseudo-labelling module (Sect.~\ref{sec:vlbc}). Finally, our method requires a generator with an editable space, pre-trained on data where the attributes to be manipulated are well-represented.

\section{Conclusion}
    In this paper, we presented a novel framework for controlling the bias in facial datasets leveraging a \textit{pre-trained} and fixed diffusion model. We built on ContraCLIP~\cite{tzelepis2022contraclip} in order to find meaningful natural language driven generative paths in the semantic space of DiffAE~\cite{preechakul2021diffusion}, which we then applied to augment a given dataset with respect to under-represented protected characteristics (e.g., black people), making it fairer for downstream tasks. The proposed bias mitigation method (VLBC-) is able to counteract the bias learnt from a downstream model, while preserving accuracy and showing competitive results against existing SOTA works~\cite{wang2020towards, nam2020learning, jung2022learning}. Additionally, VLBC- exhibits consistency across multiple settings, a trait missing in concurrent works~\cite{nam2020learning, jung2022learning}. Finally, we showed that the proposed framework, besides mitigating bias, is also capable of increasing it (VLBC+), providing full control over bias towards specific attributes. As an interesting future direction, we consider the extension of our method beyond binary classification downstream tasks.

{\setlength{\parindent}{0cm}
\textbf{Acknowledgments:} This work was supported by the MUR PNRR project FAIR (PE00000013) funded by the NextGenerationEU, the EU project ELIAS (No. 101120237), and the Junior Fellows Program of EU H2020 project AI4Media (No. 951911).
}

{\small
\bibliographystyle{ieee_fullname}
\bibliography{egbib}
}

\end{document}